\documentclass[conference]{IEEEtran}
\IEEEoverridecommandlockouts
\usepackage{cite}
\usepackage{amsmath,amssymb,amsfonts}
\usepackage{algorithmic}
\usepackage{graphicx}
\usepackage{textcomp}
\usepackage{xcolor}
\usepackage{hhline}
\usepackage{subcaption}
\usepackage{caption}
\usepackage{amsthm}
\usepackage{bm}
\usepackage{multirow}
\def\BibTeX{{\rm B\kern-.05em{\sc i\kern-.025em b}\kern-.08em
    T\kern-.1667em\lower.7ex\hbox{E}\kern-.125emX}}
\begin{document}

\title{Self-explaining Hierarchical Model for Intraoperative Time Series}

\author{
\IEEEauthorblockN{Dingwen Li\IEEEauthorrefmark{1}, Bing Xue\IEEEauthorrefmark{1}, Christopher King\IEEEauthorrefmark{2}, Bradley Fritz\IEEEauthorrefmark{2}, Michael Avidan\IEEEauthorrefmark{2}, Joanna Abraham\IEEEauthorrefmark{2}, Chenyang Lu\IEEEauthorrefmark{1}\thanks{The last author is the corresponding author.}}
\IEEEauthorblockA{\IEEEauthorrefmark{1}McKelvey School of Engineering, Washington University in St. Louis}
\IEEEauthorblockA{\IEEEauthorrefmark{2}School of Medicine, Washington University in St. Louis
\\\{dingwenli, xuebing, christopherking, bafritz, avidanm, joannaa, lu\}@wustl.edu}
}

\maketitle

\begin{abstract}
Major postoperative complications are devastating to surgical patients. Some of these complications are potentially preventable via early predictions based on intraoperative data. However, intraoperative data comprise long and fine-grained multivariate time series, prohibiting the effective learning of accurate models. The large gaps associated with clinical events and protocols are usually ignored. Moreover, deep models generally lack transparency. Nevertheless, the interpretability is crucial to assist clinicians in planning for and delivering postoperative care and timely interventions. Towards this end, we propose a hierarchical model combining the strength of both attention and recurrent models for intraoperative time series. We further develop an explanation module for the hierarchical model to interpret the predictions by providing contributions of intraoperative data in a fine-grained manner. Experiments on a large dataset of 111,888 surgeries with multiple outcomes and an external high-resolution ICU dataset show that our model can achieve strong predictive performance (i.e., high accuracy) and offer robust interpretations (i.e., high transparency) for predicted outcomes based on intraoperative time series.
\end{abstract}

\section{Introduction}
\begin{figure}[!ht]
        \centering
		\includegraphics[scale=0.38]{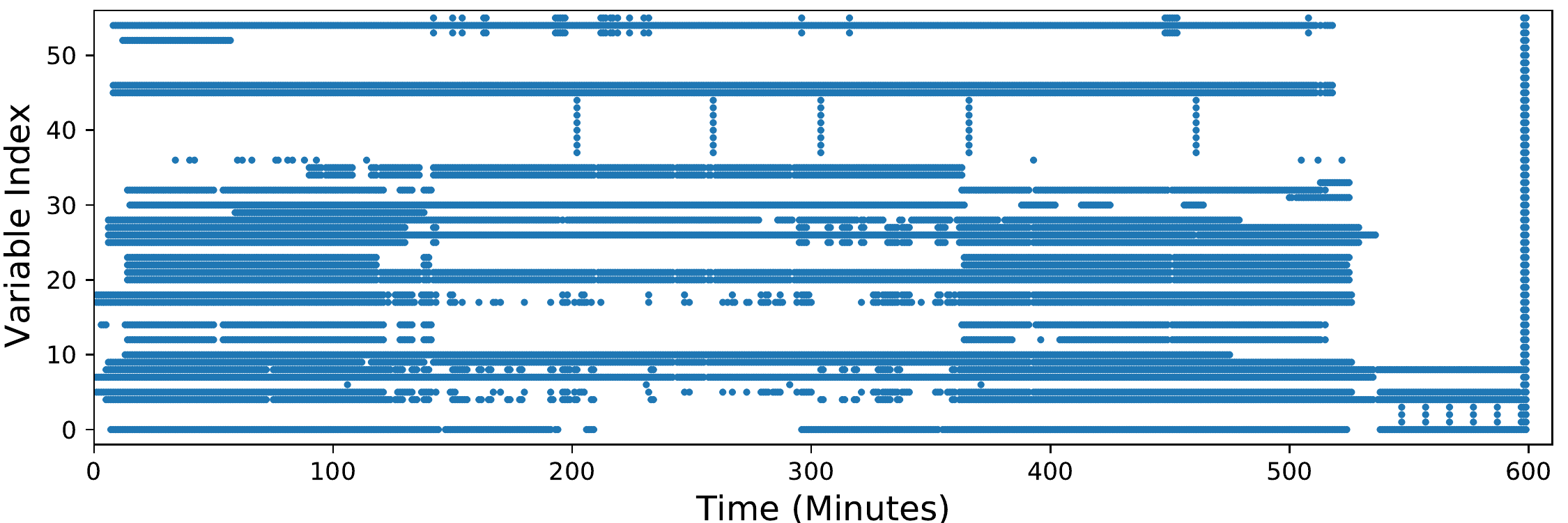}
        \caption{An example of long intraoperative time series with large gaps. Blue dots represent measurements collected from a surgical case.}\label{fig:var_vis}
\end{figure}

Major postoperative complications are devastating to surgical patients with increased mortality risk, need for care, length of postoperative hospital stay and costs of care~\cite{xue2021,TEVIS2013}. With massive electronic intraoperative data and recent advances in machine learning, some of these complications are potentially preventable via early predictions~\cite{Weller2018}.  Intraoperative data comprise long and fine-grained multivariate time series, such as vital signs and medications.  For example, Figure~\ref{fig:var_vis} visualizes the intraoperative data collected for a surgical case, which lasts for longer than 600 minutes at a sampling rate up to one per minute. 
Furthermore, there are large gaps consisting of many consecutive missing values. These gaps are often associated with the surgical procedure or clinical events that require different variables to be monitored at different stages of the surgery.  

It is challenging to learn effective representations from the long time series as modeling latent patterns from high temporal complexity is hard. Recurrent neural networks (RNNs) 
have been widely used for learning dynamics from the sequential input. However, hundreds of time steps prohibit RNNs from learning accurate representation, due to the vanishing gradient issue. A common approach to tackle the long input sequence for RNNs is to add convolutional layers before the recurrent layers~\cite{raim2018,Tan2019}. However, the introduction of a stack of convolutional layers before recurrent layers increases the complexity leading to vanishing gradients. 
Another alternative to RNN is the attention approach. Attention, e.g., Transformer~\cite{transformer2017}, can capture salient data patterns by skipping recurrent connections, thus avoiding the vanishing gradient issue brought by the long-term dependencies. Nevertheless, pure attention models cannot exploit long-term progression patterns of intraoperative time series, which are informative given the nature of physiological changes during the operation.

Another challenge of learning with intraoperative time series is associated with the large data gaps commonly observed in intraoperative time series. While imputation techniques have been investigated extensively to estimate missing values, they cannot preserve the information carried by the data gaps. The information may be exploited by predictive models given their potential association with surgical procedures and clinical events.

Furthermore, the interpretability of machine learning models, as explaining which and how input variables contribute to the predictive outcomes, is crucial to the clinicians. A good explanation helps clinicians understand the risk factors, thus knowing how to plan for and deliver postoperative care and timely interventions. Despite the invention of model-agnostic explanation methods~\cite{lime2016,shap2017}, attribution methods tailored for deep models~\cite{intgrad2017,deeplift2017}, and self-explaining models~\cite{raim2018,hitanet2020,Lim2021,Hsieh2021,stam2021,senn2018}, it remains challenging to generate accurate explanations identifying important data segments in fine-grained time series that can benefit clinicians and medical research.  

In this paper, we propose a novel \textbf{S}elf-\textbf{E}xplaining \textbf{H}ierarchical \textbf{M}odel (\textbf{SEHM}) to learn representations from long multivariate time series with large gaps and generate accurate explanations pinpointing the clinically meaningful data points. The hierarchical model comprises a kernelized local attention and a recurrent layer, which effectively 1) captures local patterns while reducing the size of the intermediate representations via the attention and 2) learns long-term progression dynamics via the recurrent module. To make the model end-to-end interpretable, we design a linear approximating network parallel to the recurrent module that models the behavior of a recurrent module locally. 

We evaluate SEHM on an extensive dataset from a major research hospital with experiments on predicting three postoperative complications and High time Resolution ICU Dataset (HiRID)~\cite{hirid2020} on predicting circulatory failure. In the evaluation, we show SEHM outperforms other state-of-the-art models in predictive performance. We also demonstrate the proposed model achieves better computational efficiency, which would be an advantage in supporting clinical decisions for perioperative care. We evaluate the model interpretability through both quantitative evaluation on the dataset and clinician reviews of exemplar surgical cases. Results suggest the advantage of SEHM over existing model interpretation approaches in identifying data samples in the input time series with potential clinical importance.

The main contributions of our work are four-fold: (1) we present a novel hierarchical model with kernelized local attention to effectively learn representations from intraoperative time series; (2) we significantly improve the computational efficiency of the hierarchical model by reducing the size of intermediate learned representation to the recurrent layer; (3) we propose a linear approximating network to model the behavior of the RNN module, which can be integrated with the kernelized local attention to establish an end-to-end interpretable model with three theoretical properties guaranteed; (4) we evaluate SEHM with experiments from both computational as well as clinical perspectives and demonstrate the end-to-end interpretability of SEHM on large datasets with multiple predictive outcomes. 

\section{Related Work}
In this section, we review the literature from three perspectives: A) models designed for handling long sequential data, B) techniques for handling missing values in time series, and C) model interpretation techniques and self-explaining models.

Traditional RNN models are widely used for learning with sequential data. However, they are ineffective when dealing with long sequential data due to the vanishing gradient issue and computation cost of recurrent operations. Temporal convolutional network (TCN), e.g., WaveNet~\cite{tcn2016}, can capture long-range temporal dependencies via dilated causal convolutions. A more recent work suggests that TCN outperforms RNN in various prediction problems based on sequential data, particularly when the input sequences are long~\cite{tcnbai2018}. However, TCN models rely on deep hierarchy to ensure the causal convolutions and thus achieve large receptive fields. Deep hierarchy, namely a large stack of layers, incurs significant computation cost for inference at run time. Efficient attention models adapted from Transformer~\cite{transformer2017} have been proposed recently for learning representations from long sequential data, which mainly focus on replacing the quadratic dot-product attention calculation with more efficient operations~\cite{informer2021,performer2021}. In this work, SEHM builds on previous insights and introduces a hierarchical model that integrates kernelized local attention and RNN. Kernelized local attention captures important local patterns and reduces the size of intermediate representation, while the higher-level RNN model learns long-term dynamics. As a result, SEHM can achieve better predictive performance and computational efficiency when learning and inferring from long multivariate intraoperative time series.

Missing values are prevalent in clinical data. They provide both challenges and information for predicting clinical outcomes. Standalone imputation models~\cite{mice2011,knnimp2016,e2gan2019} impute missing values at the preprocessing stage. However, imputation in the preprocessing stage prevents models from exploiting predictive information associated with gaps. Recently, researchers introduced imputation approaches that can be integrated with predictive models in an end-to-end manner. RNN-based imputation models, such as GRU-D~\cite{grud2018} and BRITS~\cite{brits2018}, demonstrate better performance when learning on sequential data with missing values. However, the recurrent nature of these models makes it difficult to perform imputation and predictions on long sequences. An alternative to imputation is to treat data with missing values as irregularly sampled time series. In this direction, models like multi-task Gaussian process RNN (MGP-RNN)~\cite{mgprnn2017} and neural ordinal differential equations (ODE) based RNN~\cite{latentode2019} have been proposed to accommodate the irregularity by creating evenly-sampled latent values. However, these models are computationally prohibitive for long sequences as they either operate with a very large covariance matrix or forward intermediate values to an ODE solver numerous times. We note that the aforementioned imputation approaches are not suitable for handling large gaps in time series that are common in intraoperative data, because uncertainty in missing values grows with the time elapsed from the last observed data.  Moreover, the large gaps in intraoperative time series may reflect information of the surgery. In the design of kernelized local attention, we overcome this issue by taking advantage of the characteristics of locality and using 0s to represent the missing values. This design can encode the gap information, which helps capture clinical information associated with the gaps.

Several approaches have been proposed for interpreting the predictions made by machine learning models, including model-agnostic approaches and feature attribution approaches designed for deep models. Model-agnostic explanation approaches, such as LIME~\cite{lime2016} and  SHAP~\cite{shap2017}, provide general frameworks for different models while treating them as black-box models. There are also feature attribution approaches designed for interpreting neural networks~\cite{intgrad2017,lrp2015,ancona2018towards,deeplift2017}. Deep models are not always black boxes. When properly designed attention models can be explainable by itself. Self-explaining models allow predictions be interpreted using attention matrices directly~\cite{raim2018,Lim2021,hitanet2020,Hsieh2021,stam2021}. In particular, RAIM~\cite{raim2018}, HiTANet~\cite{hitanet2020} and STAM~\cite{stam2021} are self-explaining attention models designed for interpreting clinical outcome predictions.  Alvarez-Melis et al. propose self-explaining neural networks (SENN)~\cite{senn2018} that have relevance parametrizers for interpretability, which can be optimized jointly with the classification objective. However, these self-explaining deep models are not interpretable \textit{end-to-end}. In the aforementioned models, the explanations are generated for concept bases~\cite{senn2018} or intermediate representations~\cite{raim2018,hitanet2020,stam2021}, instead of raw inputs. The concepts bases~\cite{senn2018} or intermediate representations~\cite{raim2018,hitanet2020,stam2021} do not necessarily reflect the contributions of raw inputs to the predictive outcomes due to the non-linear transformation from the raw inputs to concepts bases~\cite{senn2018} or intermediate representations~\cite{raim2018,hitanet2020,stam2021}.

In contrast, our SEHM is specifically designed to provide end-to-end interpretability by generating decomposed data contribution matrices associated with raw inputs in a linear way. SEHM also comes with theoretical properties guaranteeing the quality of interpretability, which are not covered by the existing self-explaining models. We note that end-to-end interpretability is crucial for clinical applications as clinicians usually need to review the original clinical data to interpret the predictions.

\section{Self-Explaining Hierarchical Model}
\begin{figure}[!ht]
        \centering
		\includegraphics[scale=0.38]{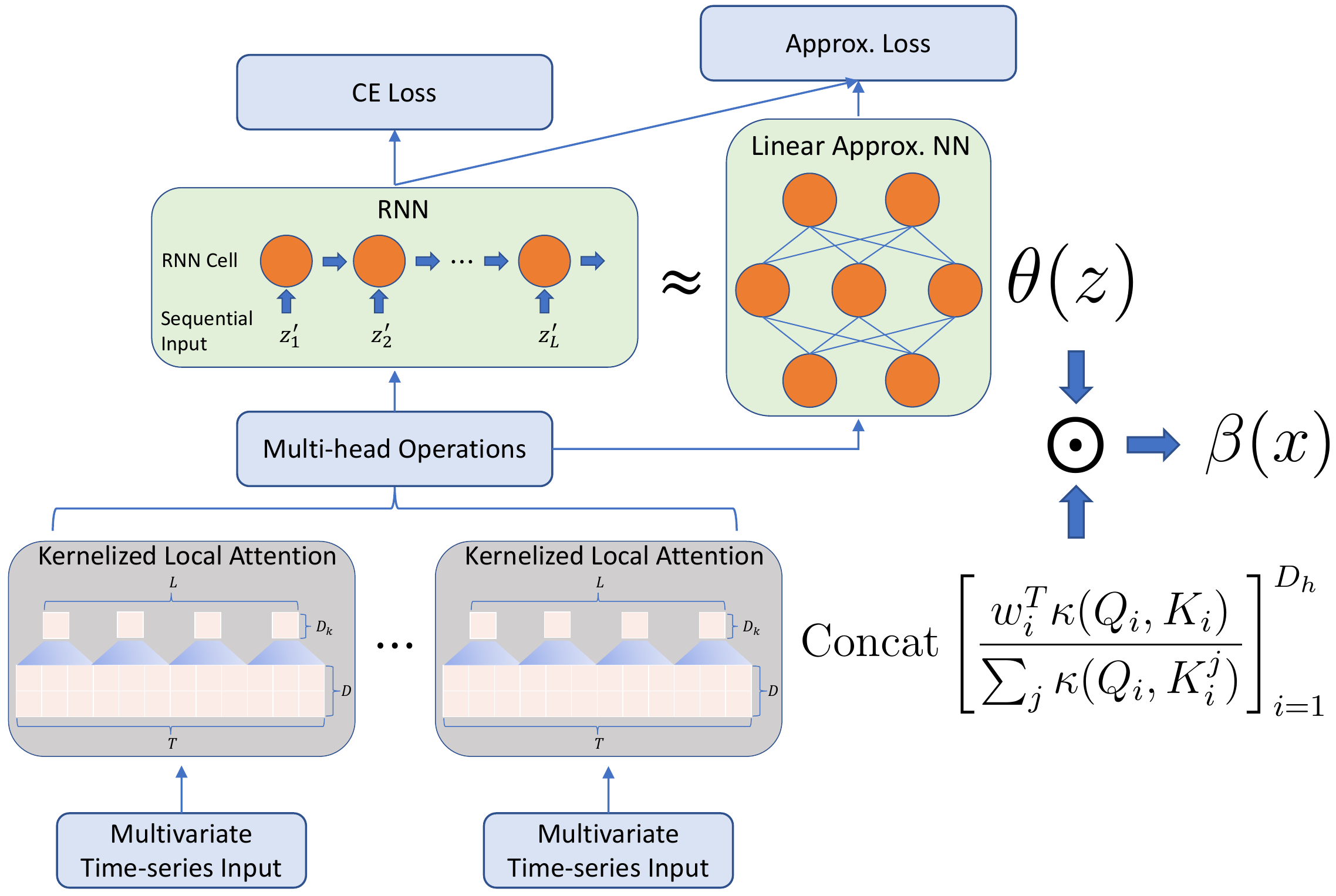}
        \caption{The overview of Self-explaining Hierarchical Model (SEHM) with multi-head kernelized local attention and linear approximating network}\label{fig:linear_approx}
\end{figure}

SEHM comprises three key components: 1) \textit{kernelized local attention} that captures important local patterns, preserves information about the data gaps, and reduces the computational complexity; 2) a \textit{recurrent layer} that learns the long-term dynamics; 3) a \textit{linear approximating network} for interpreting the recurrent layer locally. As shown in Figure~\ref{fig:linear_approx}, the input high-resolution time series firstly go through multiple kernelized local attention modules in parallel, the outputs of which are concatenated as an intermediate output via multi-head operations. The intermediate output is used as input to both recurrent layers and the linear approximating network. The cross-entropy loss and approximation loss are used for classification task and interpreting RNN, respectively.

\subsection{Kernelized Local Attention}
High-resolution clinical time series, such as intraoperative time series, usually have a length of over one hundred minutes. Such long sequences are prohibitive to traditional deep models, e.g. recurrent neural networks and attention mechanism, due to the computational complexity and vanishing gradient problem. In order to effectively and efficiently learn useful representations from the high-resolution clinical time series, we propose a kernelized local attention with the ability of exploiting short-term patterns in a temporal neighborhood via the locality structure and significantly reducing the dot-product attention's notorious quadratic complexity to linear via kernelization.

Assume we have a two-dimensional multivariate time-series input $x \in \mathbb{R}^{T\times D}$. In order to calculate the attention out of the neighbors, we reshape the input to three-dimensional tensor $\Tilde{x} \in \mathbb{R}^{L\times C \times D}$, such that $T=L\times C$. This essentially enforces the attention weights attending to the neighbors with size $C$ and outputs $L$ computed attentions. The benefits are two-fold. On one hand, self-attention allows each time step to interact with all its neighbors, which significantly reduces the information decay compared to RNN models. On the other hand, the attention weights can be associated with each neighboring time step, which allows direct interpretation on which time steps contribute most to the final outcomes. The attention matrix can be formulated as a positive-definite kernel $\kappa(q_i, k_j)$, such that $q_i$ and $k_j$ are the $i$-th vector in the query matrix and $j$-th vector in key matrix calculated from the localized expression of input $\Tilde{x}$. We define the kernelized attention as an expectation over an inner product of a randomized feature map $\phi: \mathbb{R}^D \rightarrow \mathbb{R}^R_+$ as $R>0$:
\begin{align}
    \kappa(q_i, k_j) = \mathbb{E}_{\omega\sim\mathcal{D}}[\phi(q_i)^T\phi(k_j)]
\end{align}
where $\mathcal{D}$ is a distribution from which $\omega$ is sampled i.i.d.
Thus the attention can be formulated as a weighted sum over the latent dimension (usually the temporal dimension):
\begin{align}
    a_i = \sum\limits_{j=1}^C\frac{\kappa(q_i, k_j)}{\sum_{j'=1}^C\kappa(q_i, k_{j'})}v_j = \frac{\mathbb{E}[\phi(q_i)^T\sum_{j=1}^C\phi(k_j)v_j]}{\mathbb{E}[\phi(q_i)^T\sum_{j'=1}^C\phi(k_{j'})]}.
    \label{eq:kernelattn} 
\end{align}
After reordering products and reusing $\sum_{j=1}^C\phi(k_j)v_j$ and $\sum_{j'=1}^C\phi(k_{j'})$ for each $i$, the time and memory complexity can be reduced to $O(C)$~\cite{kattn2021,performer2021}. Based on the kernel view, the Transformer's softmax function of $Q^TK$ can be approximated by kernel functions of randomized feature maps~\cite{kattn2021,performer2021}. In particular, the kernel function in Eq.(\ref{eq:kernelattn}) unbiasedly approximates the exponential of the dot product in softmax attention by drawing feature vectors from a zero-mean Gaussian distribution $\omega\sim\mathcal{N}(0, I_D)$
\begin{equation}
    \begin{split}
        \exp(q_i^Tk_j) &= 
        \mathbb{E}_{\omega\sim\mathcal{N}(0, I_D)}[\phi(q_i)^T\phi(k_j)], \\
        \text{s.t.}~\phi(z) &= \exp(\omega^Tz -\frac{\|z\|^2}{2}),~z=q_i~\text{or}~k_j.
    \end{split}
\end{equation}
This admits the decomposition of dot-then-exponential, which enables the reordering of products and reduces the time and memory complexities to linear. When constructing random feature samples $\omega$ to be exact orthogonal, the softmax attention can be accurately approximated by having exponentially small and sharper bounds on regions where the attention values after softmax are small~\cite{performer2021}.

The attention output $A=\{a_i\}_{i=1}^C$ further shrinks to aggregate the learned information among neighbors, such that $w^TA$, where $w \in \mathbb{R}^C$ is a learnable parameterized vector. Then we have the multi-head version of above attention:
\begin{align}
    H = \text{Concat}(w_1^TA_1,...w_h^TA_h,...w_H^TA_H)W^O
    \label{eq:multiatt}
\end{align}
where each $A_h^Tw_h$ denotes the attention output of head $h$, $W^O \in \mathbb{R}^{HD \times D_o}$. The learned compact representation $H$ will be used as the input to the RNN layer. The RNN layer learns the long-term dynamics in the intraoperative time series that may be associated with the post-operative complications. 

Another issue with the intraoperative time series is the large gap of missing measurements. In contrast to imputation or generative approaches, we propose to directly use original multivariate time series $\Tilde{x}$ as the "value" component $v$ in Eq.(\ref{eq:kernelattn}) and encode missing values as zeros. Zero encoding along with the special structure of the localized attention can effectively utilize the information conveyed by missing values at no additional computational costs.

\textbf{Proposition 1} 
\textit{Zero-encoding enables the kernelized local attention to output 0 for the measurement gaps $C_g\geq2C-1$, where $C$ is the size of neighborhood.}

Assume there is a gap in one of the input variable $\Tilde{x} \in \mathbb{R}^{L\times C}$ that has a length $C_g\geq2C-1$. Since $C_g\geq2C-1$, there is always at least one row in $\Tilde{x}$ that contains all zeros. Without loss of generality, we assume the $l$-th row has all zeros, denoted as $\bm{\Tilde{x}}_l=\bm{0}$. Hence, for the attention output corresponds to the $l$-th row, we can easily verify that it is equal to 0 by 
\begin{align}
    a_l =  \sum_{j=1}^C\frac{\sum_{i=1}^Cw_i\kappa(q_{li}, k_{lj})}{\sum_{j'=1}^C\kappa(q_{li}, k_{lj'})}\Tilde{x}_{lj}=0
\end{align}
where $a_l$ is the attention output corresponding to the $l$-th row vector. This design enables the attention to capture gaps that are large than $2C-1$ and preserve the information of gap in the attention output. The actual neighborhood size can be determined via cross validation.

\subsection{Self-explaining Model with Linear Approximation for RNN}
Although the kernelized local attention is explainable by itself, the hierarchical model, which consists of the attention and recurrent layers, is not end-to-end explainable due to the lack of transparency in the recurrent layers. In order to achieve end-to-end interpretability, a self-explaining linear approximation is introduced in parallel with the recurrent layers. Assume the intermediate inputs to the recurrent layers are denoted as $z$, which are interpretable bases. The linear model that is used to approximate the prediction has a form of:
\begin{align}
    g(z) = \theta(z)^T z = \sum\limits_{i=1}^{D_r}\theta_i(z)z_i
\end{align}
where $D_r$ is the dimension of $z$. We denote the whole attention layer as a function of input $x$, such that $z=h(x)$. The linear approximation can be further decomposed as a product of $\theta$ and attention parameters. The Eq.(\ref{eq:multiatt}) can be reformulated as 
\begin{align}
    g(h(x)) = \theta(h(x))^T \text{Concat}\left[\frac{w_i^T\kappa(Q_i, K_i)}{\sum_{j}\kappa(Q_i, K_i^j)}\right]_{i=1}^{D_h} \Tilde{x}
\end{align}
where $\text{Concat}[\cdot]_{i=1}^{D_h}$ denotes the concatenation of attention parameters over all the heads.
The multiplied parameters can be treated as a whole denoted by $\beta(x)$, which is a function of model input $x$:
\begin{align}
    \beta(x) = \theta(h(x))^T \text{Concat}\left[\frac{w_i^T\kappa(Q_i, K_i)}{\sum_{j}\kappa(Q_i, K_i^j)}\right]_{i=1}^{D_h}.
\end{align}
The model architecture combining kernelized local attention and the linear approximating network is shown as Figure~\ref{fig:linear_approx}.

For simplicity, in the following context we use $g(z)$ to represent the explanation model, such that
\begin{align}
    f(z) \approx g(z) = \theta(z) z
\end{align}
The goal is to make the approximated output $g(z)$ close to the actual probabilistic output $f(z)$. However, the linear approximation cannot be generalized well in a global perspective. Hence, we seek to find an accurate linear approximation locally to the input that needs to be explained. Other than the accurate approximation, we also want the explanation model to be robust against local perturbation.
If $g(z)$ is differentiable at $z$, by product rule, the gradient of $g(z)$ can be decomposed as 
\begin{align}
    \nabla_z g(z) = \theta(z)J + \nabla_z\theta(z)z \label{eq:prodrule}
\end{align}
where $J$ is an all-one matrix. In order to make $g(z)$ locally behave like a linear function and be close to the real probabilistic output $f(z)$, $\theta(z)J$ should approximate the gradient of $f(z)$, e.g., $\nabla_z f(z)$, and the second term in Eq. (\ref{eq:prodrule}), e.g., $\nabla_z\theta(z)$, should approach $\bm{0}$. With these goals, we propose a loss $\mathcal{L}_{\theta}$ to ensure the local linearity as well as stability 
\begin{align}
    \mathcal{L}_{\theta} = \| \theta(z)J - \nabla_z f(z) \|_2 + \lambda \| \nabla_z \theta(z) \|_1 \label{eq:org_loss}
\end{align}
where $\lambda$ is a coefficient balancing the two objectives. The first norm is used to enforce local linearity of the linear approximating network and the second norm is used to ensure the local approximating accuracy. However, this loss is hard to optimize in practice, since $\nabla_z \theta(z)$ has to be calculated in the loss function. In the following content, we will introduce a proposition deriving the upper bound of the aforementioned loss $\mathcal{L}_{\theta}$. This upper bound will be a surrogate loss that can be calculated efficiently with the same goal of achieving local linearity and approximating accuracy.

\textbf{Proposition 2} 
\textit{For any Multi-Layer Perceptron (MLP) implementing $\theta(z)$ with 1-Lipschitz activation functions (e.g., ReLU,
Leaky ReLU, SoftPlus, Tanh, Sigmoid, ArcTan or Softsign)~\cite{lip2018}, the upper bound of Eq. (\ref{eq:org_loss}) is}
\begin{align}
    \hat{\mathcal{L}}_{\theta} = \| \theta(z)J - \nabla_z f(z) \|_2 + \lambda \sqrt{d} \prod_{k=1}^K\|W_k\|_2\label{eq:upperbound}
\end{align}
\textit{where $W_k$ is the parameter of the k-th layer in the MLP implementing $\theta(z)$.}
\begin{proof}
From L1-L2 norm inequality, we have 
\begin{align}
    \| \nabla_z \theta(z) \|_1 \leq \sqrt{d} \| \nabla_z \theta(z) \|_2. \label{eq:init}
\end{align}
Without loss of generality, we assume that the linear approximation parameter $\theta(z)$ is realized by a nested Multi-Layer Perceptron (MLP) with $\theta_k=a_k(g_k(\theta_{k-1}))$, where $g_k$ is the k-th layer perceptron, $a_k$ is the k-th 1-Lipschitz activation function, $\theta_{k-1}$ is the output from the last preceding layer. The k-th layer perceptron takes an affine transformation on the input data, such that $g_k(\theta_{k-1}) = W_k\theta_{k-1} + b_k$. The chain rule implies that the gradient of $\theta(z)=\theta_K$ can be derived as
\begin{align}
    \nabla_z \theta(z) = a_K'g_K'\nabla \theta_{K-1}.
\end{align}
where $a_k'$ and $g_k'$ represent the Jacobian matrices, $K$ denotes the last layer in MLP. Take the 2-norm of both sides, then we get
\begin{align}
     \| \nabla_z \theta(z) \|_2 &= \| a_K'g_K'\nabla \theta_{K-1} \|_2 \leq  \\
     \leq \| a_K' \|_2 \| g_K'\nabla \theta_{K-1} \|_2
     &\leq \| a_K' \|_2 \| g_K' \|_2 \| \nabla \theta_{K-1} \|_2
\end{align}
where the norm for matrices is the induced 2-norm, $\| \nabla \theta_{K-1} \|_2$ can be further expanded via chain rule until reaching the input $z$. Since $\{a_k\}_{k=1}^K$ functions are all 1-Lipschitz activation functions, it implies that $\|a_k'\| \leq 1$. Each layer in MLP is an affine transformation, which yields the magnitude of $g_k'$ to be $\| W_k\|_2$. Thus, we have 
\begin{align}
    \|\nabla_z \theta(z)\|_2 \leq \prod_{k=1}^K\|W_k\|_2 \label{eq:gradbound}
\end{align}
assuming $g_k$ is an affine function and $a_k$ is a 1-Lipschitz activation function. With Eq. (\ref{eq:init}) and Eq. (\ref{eq:gradbound}), we obtain an upper bound $\hat{\mathcal{L}}_{\theta}$ of the original loss $\mathcal{L}_{\theta}$, such that
\begin{align}
    \mathcal{L}_{\theta} = \| \theta(z)J - \nabla_h f(z) \|_2 + \lambda \| \nabla_z \theta(z) \|_1 \\ 
    \leq \| \theta(z)J - \nabla_z f(z) \|_2 + \lambda \sqrt{d} \prod_{k=1}^K\|W_k\|_2 = \hat{\mathcal{L}}_{\theta}
\end{align}
\end{proof}

To model the local behavior of the predictive model, we randomly sample instances around $z$ uniformly within a small distance. Thus, we obtain a perturbed set of $z'\in \mathbb{Z}$, which is used for approximating $f(z)$ locally. With the derived upper bound Eq. (\ref{eq:upperbound}), we have the overall objective function for the self-explaining linear approximation:
\begin{align}
    \mathcal{L} = \sum\limits_{z'\in \mathbb{Z}}\hat{\mathcal{L}}_{\theta}(z') + \lambda_r \mathcal{R}_1
\end{align}
where $\mathcal{R}_1$ is the L1 regularization on the parameterized neural network $\theta(z)$ to enforce sparse and disentangled $\theta(z)$ associated with $z$.

The proposed self-explaining linear approximation comes with three properties.

\textbf{Property 1 (Additive Attribute Model)}
\begin{align}
    f(z) \approx g(z) = \sum\limits_{i=1}^K\theta_iz_i
\end{align}
\textit{(1) The explanation model isolates the effect of each input variable. (2) The effect of each input can be directly added to produce the final output. (3) The sign and magnitude of $\theta$ can be interpreted as the input contribution to the predicted outcome.}

\textbf{Property 2 (Dummy)}
\textit{A variable $i$ that does not have any contribution to the output should be assigned with $\theta_i=0$.} 

This property can be verified by the first part of the loss function $\| \theta(z) - \nabla_z f(z) \|_2$, which enforces $\theta(z_i) = \partial f(z)/ \partial z_i$. On the other hand, a variable $i$ that does not have any contribution to $f(z)$ is equivalent to $\partial f(z)/ \partial z_i = 0$, which means that no matter how $z_i$ changes $f(z)$ stays the same.

\textbf{Property 3 (Locally Bounded)}
\textit{For every $z_0$ and its corresponding explainable coefficient $\theta(z_0)$, there exists $\delta>0$ and $L \in \mathbb{R}$ such that $\|z-z_0\|_2<\delta$ implies $\|\theta(z)-\theta(z_0)\|\leq L \|z-z_0\|_2$.}

To ensure the explanation $\theta(z_0)$ is locally bounded, one has to verify that the gradient of the explanation $\nabla_{z=z_0}\theta(z)$ is bounded at $z_0$. This can be enforced by Eq. (\ref{eq:gradbound}), which derives an upper bound for $\|\nabla_{z}\theta(z)\|_2$. 

The self-explaining linear approximating network can be trained either with the classification loss or separately depending on the computational resource available on the machine that is used to perform inference. Combined with the attention weights, we have an end-to-end explanation model that directly quantifies the contribution of each input data point to the predicted outcome.

\section{Experimental Evaluation}
We evaluate SEHM from three perspectives: 1) predictive performance, 2) computational efficiency, and 3) interpretability. The experiments were conducted on a large dataset collected from 111,888 operations performed on adults at Barnes Jewish Hospital from June 1, 2012, to August 31, 2016. To assess the generality of the modeling approach, we evaluated predictive performance for three types of complication including delirium, pneumonia and acute kidney injury (AKI). These complications were identified to be essential for postoperative care based on a recent stakeholder-based study with clinicians. We also performed an external evaluation on HiRID~\cite{hirid2020} to validate the generality of the hierarchical model on modeling other high-resolution clinical time series.

\subsection{Dataset and Preprocessing}
\subsubsection{Postoperative Complication Prediction}
The input data were intraoperative data comprising fine-grained multivariate time series, including vital signs (e.g., heart rate, SpO2 and blood pressure), ventilator settings (e.g., tidal volume, inspiratory pressure, and ventilation frequency) and medications (e.g., norepinephrine and phenylephrine). There were 56 time-series variables in total with a maximum sampling rate of every minute. To ensure the richness of the input information, we included all observations from 600 minutes prior to the end of surgery. Missing values were handled by either built-in imputation method or zero-encoding according to different models. The label was defined as the onset of a particular postoperative complication. Thus, we extracted exactly one example from a surgical case.

After preprocessing, we obtained three datasets for evaluating the model's performance on predicting delirium, pneumonia and AKI respectively. The delirium dataset contained 12,904 samples with a positive rate of 52.6\%, which is smaller than the other two datasets due to the availability of the delirium labels for only a fraction of the surgery cases. The pneumonia dataset contained 111,888 samples with a positive rate of 2.2\%. The AKI dataset contained 106,870 samples with a positive rate of 6.1\%.

\subsubsection{Circulatory Failure Prediction}
HiRID is a freely accessible critical care dataset with high-resolution data from 36,098 patient admissions collected between January 2008 and June 2016~\cite{hirid2020}. Clinical time series, such as heart rate, were recorded at a frequency of one measurement every 2 minutes. The task is to predict circulatory failure 8 hours prior to the first occurrence~\footnote{We used the same definition of circulatory failure that was originally proposed by~\cite{hirid2020}}. We excluded admissions that were shorter than 8 hours, resulting in 134,362 samples with a positive rate of 6.8\%. 
37 time-series variables with the overall availability $>$1\% were selected. For each admission with circulatory failure, we extracted all time-series data of these 37 variables from 16 hours to 8 hours prior to the first occurrence of circulatory failure, yielding a positive sample with a maximum of 480-minute data. For the time period from the start of the admission to the 16-th hour prior to the first occurrence, we segmented it into multiple 8-hour consecutive chunks. We applied a sliding window with a stride of 8 hours to extract data of 37 variables from each chunk, yielding negative samples. For each admission without circulatory failure, we applied the same procedure as described above to extract negative samples, except the window slid along the whole admission. Similar to the complication prediction dataset, missing values were handled by either built-in imputation method or zero-encoding according to different models. 

\subsection{Evaluation Setting}
The datasets were split as 75\% of the samples were used for model training and the rest 25\% were used for testing. Within the training set, we further designated 10\% of them as a validation set for hyperparameter tuning. For all models used in the evaluation, we tuned the batch size from a set of choices, such as 16, 32, 64, 128, 256. We also tuned the learning rate of Adam optimizer from 0.0001 to 0.01. For other hyperparameters specific to each model, we applied Bayesian optimization to select an optimal set of hyperparameters based on the validation set. Each predictive accuracy evaluation was run repeatedly for 10 times. The computation speed evaluations were deployed on Nvidia GeForce 3090 GPU and Intel(R) Core(TM) i9-10850K CPU @ 3.60GHz CPU. To ensure fairness the sizes of model are particularly controlled to be similar during the computation speed comparison, so the results purely reflect the speed of different techniques. The code is available\footnote{https://github.com/WU-CPSL/sehm}.

\subsection{Predictive Performance Benchmark}
In this experiment, we evaluate the predictive performance of SEHM in comparison to a set of existing models including state-of-the-art models designed for long multivariate time series. We use the the area under the receiver operating characteristic curve (AUROC) and the area under the precision recall curve (AUPRC) as performance metrics. 

The models included in our performance evaluation can be classified into three categories. The first category includes RNN variants for sequential data.
\begin{itemize}
    \item \textbf{LSTM/GRU}: RNN models trained on the raw input.
    \item \textbf{BRITS*}~\cite{brits2018}: a bidirectional RNN with a built-in imputation component for handling missingness in the input~\footnote{The inputs to BRITS are down-sampled by a factor of 20. The original raw inputs cause slow training and the performance is sub-optimal compared to model trained with down-sampled inputs}.
    \item \textbf{GRU-D*}~\cite{grud2018}: a GRU model with a built-in imputation component for handling missingness in the input data. The input is also down-sampled by a factor of 20.
    \item \textbf{Latent-ODE}~\cite{latentode2019}: a latent neural ordinal differential equations model for irregular time-series input. 
\end{itemize}
The second category includes Transformer-type attention models for handling sequential data.
\begin{itemize}
    \item \textbf{SAnD}~\cite{sand2018}: a deep Transformer designed for clinical outcome predictions.
    \item \textbf{Informer}~\cite{informer2021}: a computationally efficient Transformer-type model for long sequences.
    \item \textbf{Performer}~\cite{performer2021}: an efficient attention model with re-designed fast attention for long sequences.
\end{itemize}
The third category includes existing deep hierarchical models and our proposed SEHM models.
\begin{itemize}
    \item \textbf{Conv LSTM/GRU}~\cite{Tan2019}: a combination of convolutional layers and recurrent layers.
    \item \textbf{Multi-scale CNN}~\cite{cui2016}: Multiple convolutional layers with different kernel size concatenated in parallel for extracting patterns in various receptive fields.
    \item \textbf{TCN}~\cite{tcn2016,tcnbai2018}: Temporal convolutional network with dilation designed to handle long input sequence.
    \item \textbf{RAIM}~\cite{raim2018}: A recurrent attentive hierarchical model designed for multimodal patient data.
    \item \textbf{SEHM(LSTM/GRU)}: Our proposed model with a multi-head kernelized local attention layer and an RNN layer on the top.
\end{itemize}
The results of the predictive performance evaluation are shown in Table~\ref{tab:model_results_1}. 
\begin{table*}[t]
\caption {Predictive performance $mean(\sigma)$ reported for different complication prediction tasks.}  \label{tab:model_results_1} 
\setlength{\tabcolsep}{3pt}
\centering
\footnotesize
\begin{tabular} { | c | c c | c c | c c | c c | }
\hline
& \multicolumn{2}{c|}{Delirium} & \multicolumn{2}{c|}{Pneumonia} & \multicolumn{2}{c|}{AKI} & \multicolumn{2}{c|}{HIRID}\\ 
& AUROC & AUPRC & AUROC & AUPRC & AUROC & AUPRC & AUROC & AUPRC \\ \hline
LSTM & 0.7083(0.0035) & 0.7192(0.0042) & 0.8323(0.0010) & 0.0966(0.0012) & 0.7199(0.0013) & 0.1796(0.0051) & 0.8572(0.0055) & 0.3924(0.0241) \\ \hline
GRU & 0.7182(0.0013) & 0.7369(0.0025) & 0.8409(0.0017) & 0.1155(0.0033) & 0.7560(0.0021) & 0.1921(0.0013) & 0.8611(0.0035) & 0.4156(0.0222)\\ \hline
BRITS* & 0.7438(0.0010) & 0.7684(0.0016) & 0.8509(0.0018) & 0.1384(0.0023) & 0.7815(0.0006) & 0.2182(0.0023) & 0.9181(0.0015) & 0.5354(0.0080) \\ \hline
GRU-D* & 0.7386(0.0018) & 0.7605(0.0020) & 0.8510(0.0016) & 0.1349(0.0038) & 0.7698(0.0026) & 0.2100(0.0012) & 0.9068(0.0103) & 0.5143(0.0077) \\ \hline
Latent-ODE & 0.7294(0.0021) & 0.7551(0.0019) & 0.8406(0.0038) & 0.1314(0.0050) & 0.7663(0.0049) & 0.2068(0.0032) & 0.8876(0.0134) & 0.5009(0.0065) \\ 
 \hhline{|=|==|==|==|==|}
SAnD & 0.7274(0.0042)  & 0.7575(0.0052) & 0.8215(0.0053) & 0.1121(0.0032) & 0.7565(0.0056) & 0.1938(0.0073) & 0.8963(0.0074) & 0.4539(0.0053) \\ \hline
Informer & 0.7351(0.0009) & 0.7627(0.0024) & 0.8347(0.0034) & 0.1206(0.0049) & 0.7597(0.0016) & 0.1955(0.0006) & 0.9078(0.0086) & 0.5220(0.0055) \\ \hline
Performer & 0.7301(0.0033) & 0.7581(0.0025) & 0.8383(0.0056) & 0.1192(0.0044) & 0.7532(0.0015) & 0.1888(0.0049) & 0.9043(0.0044) & 0.5178(0.0057) \\
 \hhline{|=|==|==|==|==|}
Conv LSTM & 0.7392(0.0038) & 0.7647(0.0027) & 0.8450(0.0012) & 0.1358(0.0020) & 0.7576(0.0015) & 0.1976(0.0015) & 0.9118(0.0089) & 0.5328(0.0065) \\ \hline
Conv GRU & 0.7369(0.0015) & 0.7586(0.0014) & 0.8503(0.0008) & 0.1388(0.0015) & 0.7763(0.0005) & 0.2080(0.0014) & 0.9150(0.0021) & 0.5319(0.0023) \\ \hline
Multi-scale CNN & 0.7397(0.0013) & 0.7652(0.0010) & 0.8504(0.0016) & 0.1411(0.0018) & 0.7769(0.0019) & 0.2123(0.0031) & 0.8952(0.0035) & 0.4973(0.0065) \\ \hline
TCN & 0.7369(0.0013) & 0.7552(0.0019) & 0.8401(0.0023) & 0.1148(0.0064) & 0.7444(0.0010) & 0.1915(0.0015) & 0.8908(0.0117) & 0.4917(0.0022)\\ \hline
RAIM & 0.7228(0.0038) & 0.7509(0.0039) & 0.8423(0.0005) & 0.1314(0.0009) & 0.7644(0.0008) & 0.2045(0.0028) & 0.9039(0.0076) & 0.5034(0.0064) \\ 
 \hhline{|=|==|==|==|==|}
SEHM(LSTM) & 0.7565(0.0017) & 0.7789(0.0030) & 0.8587(0.0012) & 0.1496(0.0028) & 0.8086(0.0018) & \textbf{0.2380(0.0055)} & \textbf{0.9273(0.0025)} & \textbf{0.5651(0.0014)} \\ \hline
SEHM(GRU) & \textbf{0.7571(0.0015)} & \textbf{0.7795(0.0011)} & \textbf{0.8610(0.0009)} & \textbf{0.1505(0.0026)} & \textbf{0.8116(0.0024)} & 0.2378(0.0033) & 0.9265(0.0012) & 0.5628(0.0020) \\ \hline
\end{tabular}
\end{table*}
We have the following observations from Table~\ref{tab:model_results_1}. (1) Both SEHM variants outperform their vanilla RNN baselines in terms of AUROC and AUPRC. The comparison shows the advantage of using kernelized local attention to capture important local patterns and shortening the inputs to latter RNN models. (2) SEHM models have better results than BRITS, GRU-D and Latent-ODE, which suggests the proposed kernelized local attention technique with zero encoding to represent missing values is beneficial for time series with large gaps. (3) Both SEHM variants demonstrate better performance than the pure attention models, which indicates the necessity of incorporating RNN models for learning long-term dynamics. (4) When comparing SEHM models with other hierarchical models using convolution as the first layer, we observe the introduction of locality to attention is a better way of learning local patterns than convolution for intraoperative time series, as the locality association can be learned adaptively via attention. (5) The consistent results across different prediction tasks suggest that the approach may be generalizable for predicting different postoperative complications. All the aforementioned results on the comparison between SEHM and baselines are statistically significant (T test $p<0.05$).

\begin{figure*}[!ht]
     \centering
     \begin{subfigure}[b]{0.30\textwidth}
         \centering
         \includegraphics[width=\textwidth]{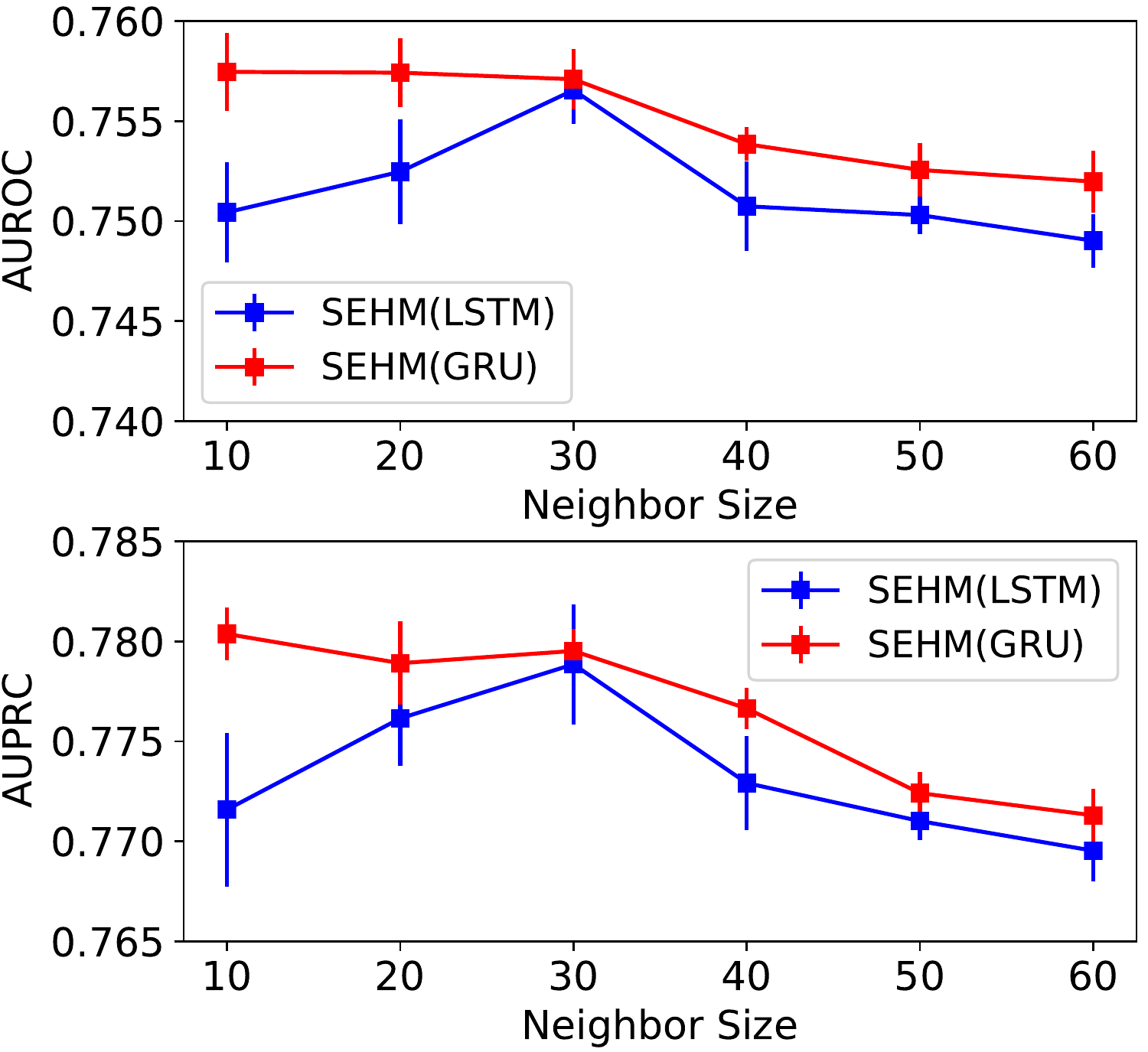}
         \caption{Delirium}
         \label{fig:del_vary}
     \end{subfigure}
     \hfill
     \begin{subfigure}[b]{0.30\textwidth}
         \centering
         \includegraphics[width=\textwidth]{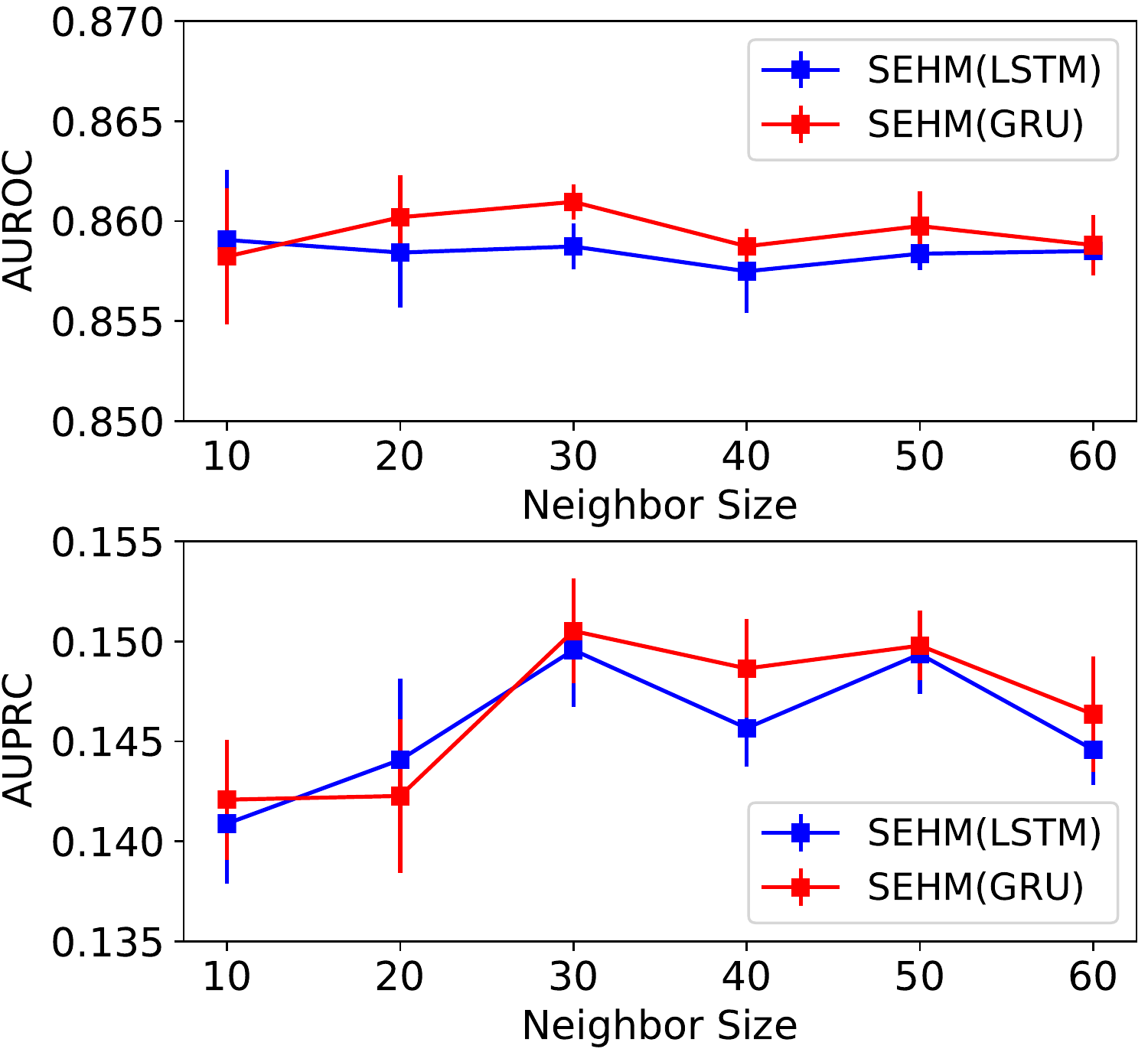}
         \caption{Pneumonia}
         \label{fig:pneu_vary}
     \end{subfigure}
     \hfill
     \begin{subfigure}[b]{0.30\textwidth}
         \centering
         \includegraphics[width=\textwidth]{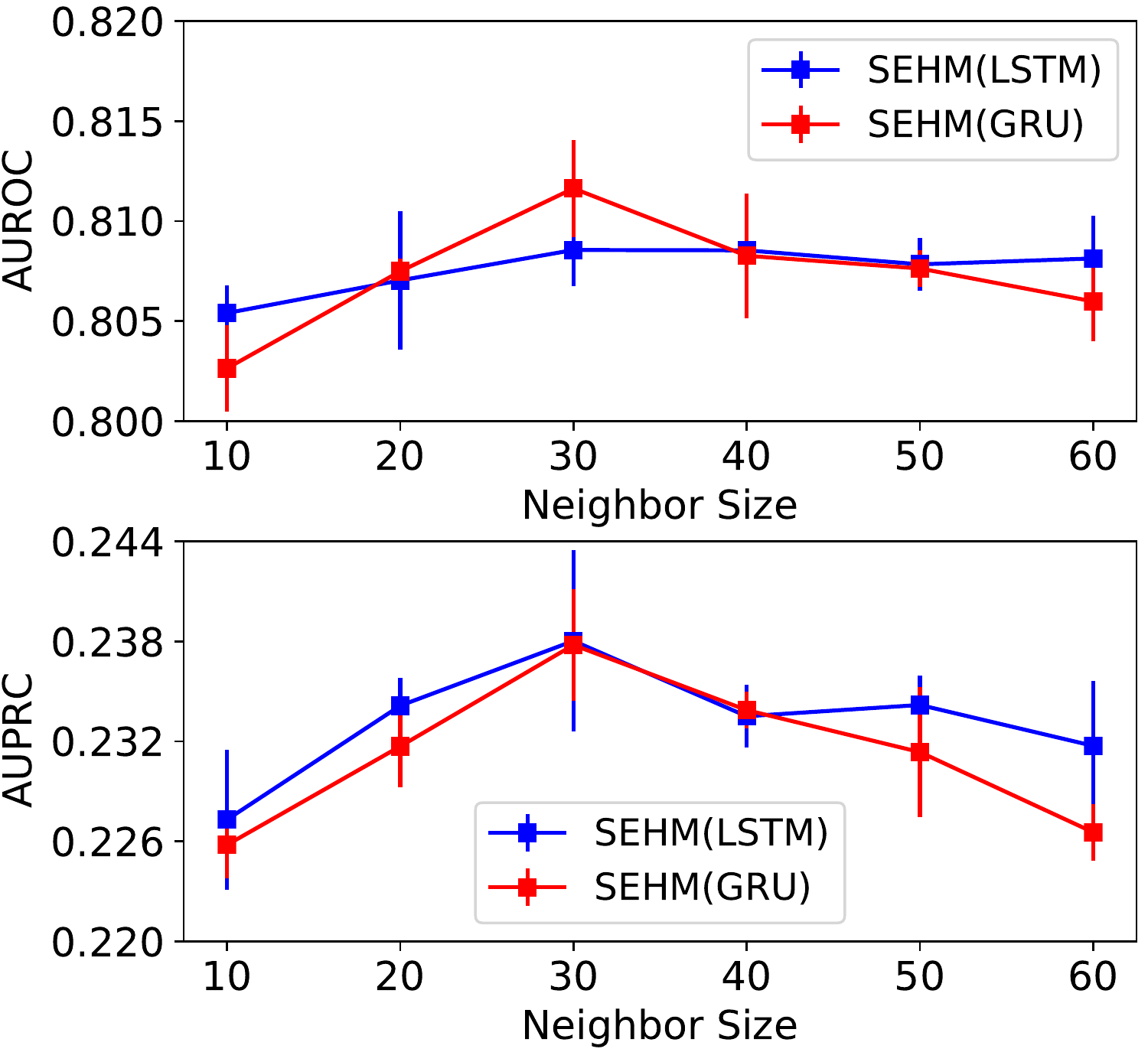}
         \caption{AKI}
         \label{fig:aki_vary}
     \end{subfigure}
        \caption{Predictive performance of SEHM models with different neighbor sizes}
        \label{fig:neighbor_size}
\end{figure*}

The second experiment is designed to explore the relation between neighbor size in kernelized local attention and predictive performance. In the experiment, we vary the neighbor size from 10 to 60, inclusive, with a stride of 10 and plot the trend of AUROC and AUPRC. The results are visualized in Figure~\ref{fig:neighbor_size} for the three complication predictions.
The optimal neighbor size is 30 for the three complication predictions, except for SEHM(GRU) trained for delirium prediction for which the optimal neighbor size is 10. We note that with all these different choices of neighbor size SEHM models are able to outperform other baselines. We also observe that the AUROC and AUPRC of SEHM(LSTM) on delirium predictions and the AUPRC of SEHM(LSTM) on pneunomia and AKI predictions change significantly with different neighbor sizes, which suggests the necessity of tuning neighbor sizes for different prediction problems.

\subsection{Ablation Study}
We performed ablation study on the delirium prediction to evaluate the effect of locality, zero-encoding and kernelization in terms of predictive performance and model inference speed. The inference speed is measured as the average time (in milliseconds) of completing a forward inference with a batch size of 64 samples. In this ablation study, our goal is to validate the effect of each technique proposed for the attention part. Thus, the overall hierarchical structure stays the same, such that the output of attention module is directly fed into the RNN module. We selected LSTM as the RNN module and it was unchanged in the ablation study. The first model in the ablation study is the pure Transformer-type attention. Then, different techniques are added to the attention part, as shown in Table~\ref{tab:ablation}.
\begin{table}[!h]
\caption {Ablation study} \label{tab:ablation} 
\centering
\footnotesize
\setlength{\tabcolsep}{3pt}
\begin{tabular} { | c c c | c c c | }
\hline
 Locality & Zero & Kernel- & AUROC & AUPRC & Time (ms)\\ 
 & encoding & ization &  &  &  \\ \hline
 & & & 0.7233(0.0034) & 0.7450(0.0019) & 37.5(3.1) \\ \hline
 & \checkmark & & 0.7342(0.0033) & 0.7542(0.0012) & 37.1(2.8) \\ \hline
 \checkmark & & & 0.7342(0.0025) & 0.7615(0.0029) & 17.9(2.5) \\ \hline
 & & \checkmark & 0.7214(0.0018) & 0.7435(0.0033) & 32.4(3.7) \\ \hline
 \checkmark & \checkmark &  & 0.7565(0.0017) & 0.7789(0.0030) & 18.5(3.1)\\ \hline
 \checkmark &  & \checkmark & 0.7398(0.0014) & 0.7651(0.0007) & 12.1(1.8) \\ \hline
 & \checkmark  & \checkmark & 0.7330(0.0023) & 0.7547(0.0028) & 33.9(2.3) \\ \hline
 \checkmark & \checkmark & \checkmark & 0.7530(0.0011) & 0.7806(0.0019) & 11.8(2.2) \\ \hline
\end{tabular}
\end{table}
The ablation study gives us following observations. (1) Locality reduces the inference time significantly while yielding better predictive results. This is because the temporal size of input to latter RNN model is reduced and local attention exploits useful information from temporal neighbors. (2) The introduction of zero encoding improves the predictive performance without additional computation overhead. (3) The kernelization further increases the model inference speed and achieves comparable predictive performance as the original softmax function.

\subsection{Computational Efficiency}
\begin{figure}[!h]
\centering
\includegraphics[scale=0.36]{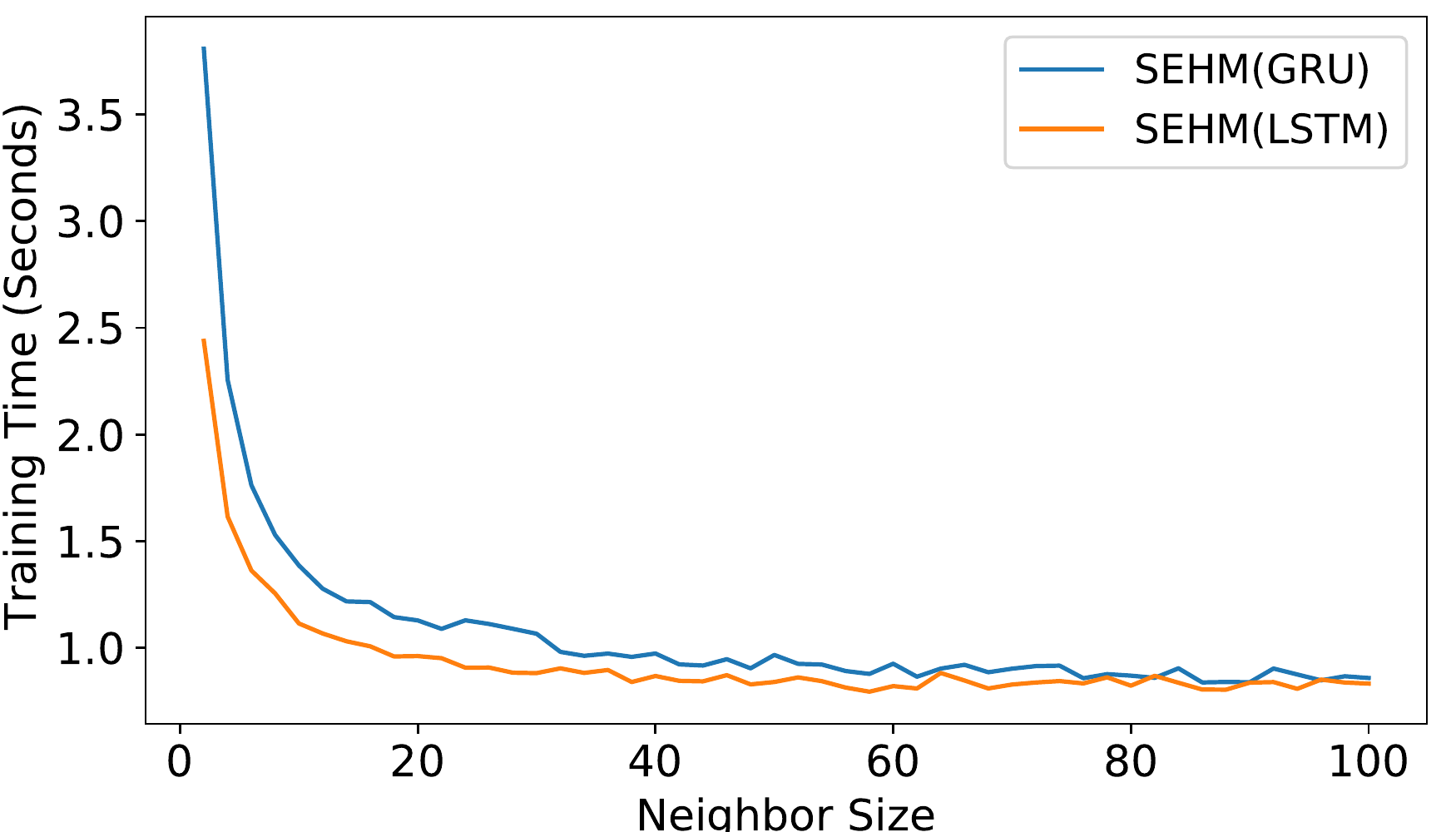}
    \caption{Training time with varying neighbor size}
    \label{fig:varying_time}
\end{figure}

We aimed to investigate the relation between model training time and the neighbor size defined in the kernelized local attention. In our empirical evaluation on the delirium subset we measured the run time in the training phase along with the varying neighbor size. The run time in the training phase is the average recorded time of executing one training epoch with a batch size of 64. As shown in Figure~\ref{fig:varying_time}, when neighbor size is less than 20, the training time drops drastically as the neighbor size increases. The gain in computational efficiency saturates when the neighbor size is greater than approximately 60. As the neighbor size increases from a very small number, the training time decreases drastically. However, when we continuously increase the neighbor size $k$, we get diminished return in reducing the training time. The overall training time should also asymptotically approach a constant number including the time needs to train parameters in the kernelized attention, which has a theoretical complexity of $O(Ld)$. This behavior can be observed and verified by the trends of actual training time as shown in Figure~\ref{fig:varying_time}. Thus, referring to the predictive results in Figure~\ref{fig:neighbor_size}, we conclude that the neighbor size should be appropriately chosen, which cannot be either too small or too large, to achieve optimal predictive performance and computational efficiency.

\subsection{Evaluations on Interpretability}
In this section, we evaluate the interpretability of the explanation generated by our model compared to the state-of-the-art model explanation approaches including model-agnostic approaches, feature attribution approaches for deep models and a self-explaining model designed for fine-grained clinical time series. The experiments include quantitative evaluations as well as case studies.

The model explanation methods used in the evaluations are:
\begin{itemize}
    \item \textbf{LIME}~\cite{lime2016}: a model-agnostic explanation method based on local linear approximation;
    \item \textbf{SHAP}~\cite{shap2017}: a model-agnostic explanation method based on assigning Shapley values to input data; both KernelSHAP and DeepSHAP are evaluated;
    \item \textbf{Integrated Gradient}~\cite{intgrad2017}: a explanation method computing the gradient of the prediction with respect to the input;
    \item \textbf{DeepLift}~\cite{deeplift2017}: a recursive explanation method attributing activation differences to the input via backpropagation;
    \item \textbf{RAIM}~\cite{raim2018}: a self-explaining deep model that uses attention matrices as model explanations.
\end{itemize}
    
\begin{table}[!h]
\caption {Quantitative evaluation of model explanation approaches $mean(\sigma)$} \label{tab:quantitative_eval} 
\centering
\footnotesize
\begin{tabular} { | c | c  c  c | }
\hline
 & Local Accuracy $\downarrow$ & Faithfulness $\uparrow$ & Stability $\downarrow$ \\ 
 & (MSE) & (AOPC@2k) & (est. Lipschitz) \\ \hline
LIME & 0.2957(0.0374) & 0.1934(0.0005) & 12.3944(0.0114) \\ \hline
KernelSHAP & 0.3241(0.0215) & 0.1141(0.0035) & 10.1523(0.3258) \\ \hline
DeepSHAP & 0.3837(0.0084) & 0.1118(0.0112) & 8.7104(0.2246) \\ \hline
Int. Grad. & 0.3178(0.0145) & 0.2749(0.0041) & 5.7964(0.1762) \\ \hline
DeepLift & 0.2648(0.0027) & 0.3321(0.0060) & 8.9637(0.2714) \\ \hline
RAIM & -- & 0.1513(0.0025) & 5.3167(0.2988) \\ \hline
SEHM & \textbf{0.2327(0.0118)} & \textbf{0.5583(0.0057)} & \textbf{3.5498(0.0811)} \\ \hline
\end{tabular}
\end{table}

\subsubsection{Quantitative Evaluation}
We propose three evaluation metrics for comparing the explanations generated by different approaches. \textit{Local accuracy} is defined as the mean square error between the aggregated explanations generated by the model explanation approach and the probabilistic outputs of the original predictive model. We note that there is no local accuracy evaluation for RAIM, since it is not an additive feature attribution method. Local accuracy reflects how accurate the summed explanation fits to the predicted output. \textit{Faithfulness} is achieved by evaluating the area over the most relevant first perturbation curve (AOPC), which assesses the ability of model assigning high values to those input variables that have the true high influence to the final predictive outcomes~\cite{Samek2017}. In our evaluation, we assess the AOPC of model explanation approaches at different cutoff points along the rank of feature importance. We also report the AOPC of top 2,000 data points ranked by the model explanation methods. \textit{Stability} is defined as the extent of changes in explanation when applying small perturbation to the input that does not change the predictive outcome. In our evaluation, we use the estimated Lipschitz continuity~\cite{senn2018} to quantify the stability of the explanation. The explanations generated with smaller estimated Lipschitz continuity should be more stable.

\begin{figure}[!h]
        \centering
		\includegraphics[scale=0.36]{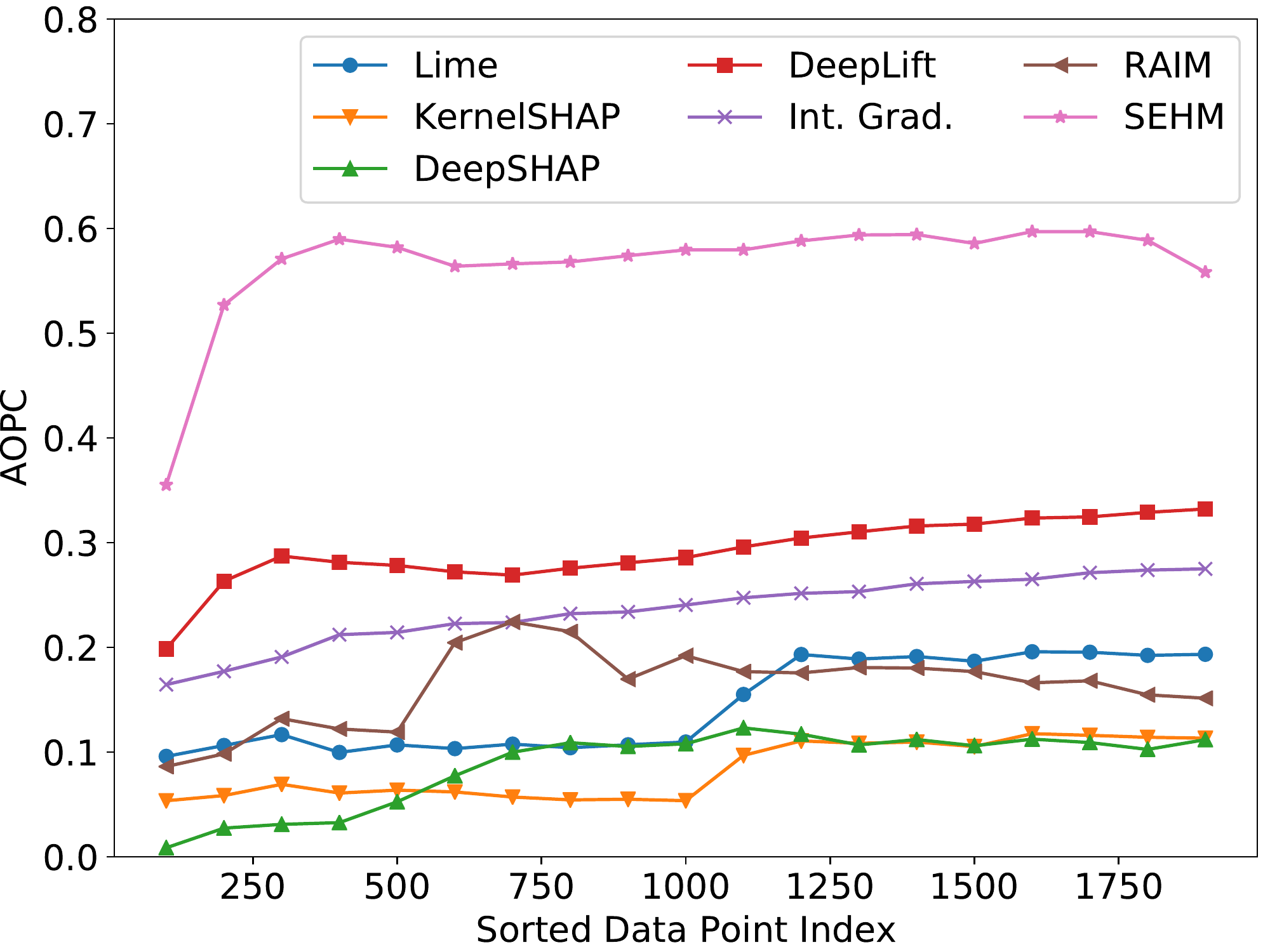}
        \caption{AOPC with increasing data points sorted by the rank of importance.}\label{fig:varying_aopc}
\end{figure}

We observe that SEHM significantly outperforms other baselines in the three quantitative evaluations ($p<0.05$), as detailed in Table~\ref{tab:quantitative_eval}. Since SEHM utilizes approximation to model the behavior of RNN, better local accuracy of SEHM can be interpreted as more accurately approximating the behavior of the RNN. The evaluation on the faithfulness at a cutoff of top 2,000 ranked data points along with a more detailed analysis in Figure~\ref{fig:varying_aopc} confirms that SEHM is better at identifying important data points in the intraoperative time series by ranking the most relevant data points correctly. This is a very promising result, since SEHM is able to provide clinicians with more faithful explanations and avoid wrong explanations that may trigger false alarms. The more stable explanations guarantee that the explanations generated for similar inputs should stay similar.

\subsubsection{Case Studies}
From a clinical perspective, a good explanation should pinpoint the regions with the most impacts on the predictions and help clinicians understand the critical risk factors. As case studies we visualize the explanations provided by SEHM, DeepLift, and KernelSHAP in two surgical cases and have an anesthesiologist specializing in perioperative care review the explanations from a clinical perspective. DeepLift and KernelSHAP are chosen as they are representatives of attribution methods and model-agnostic methods, respectively. The self-explaining models are not applied to the case studies as they cannot provide end-to-end explanations that attribute contributions to original clinical data.

\begin{figure}[!ht]
        \centering
        \subfloat[Case A]{
        \label{fig:qual_vis_86}
		\includegraphics[scale=0.088]{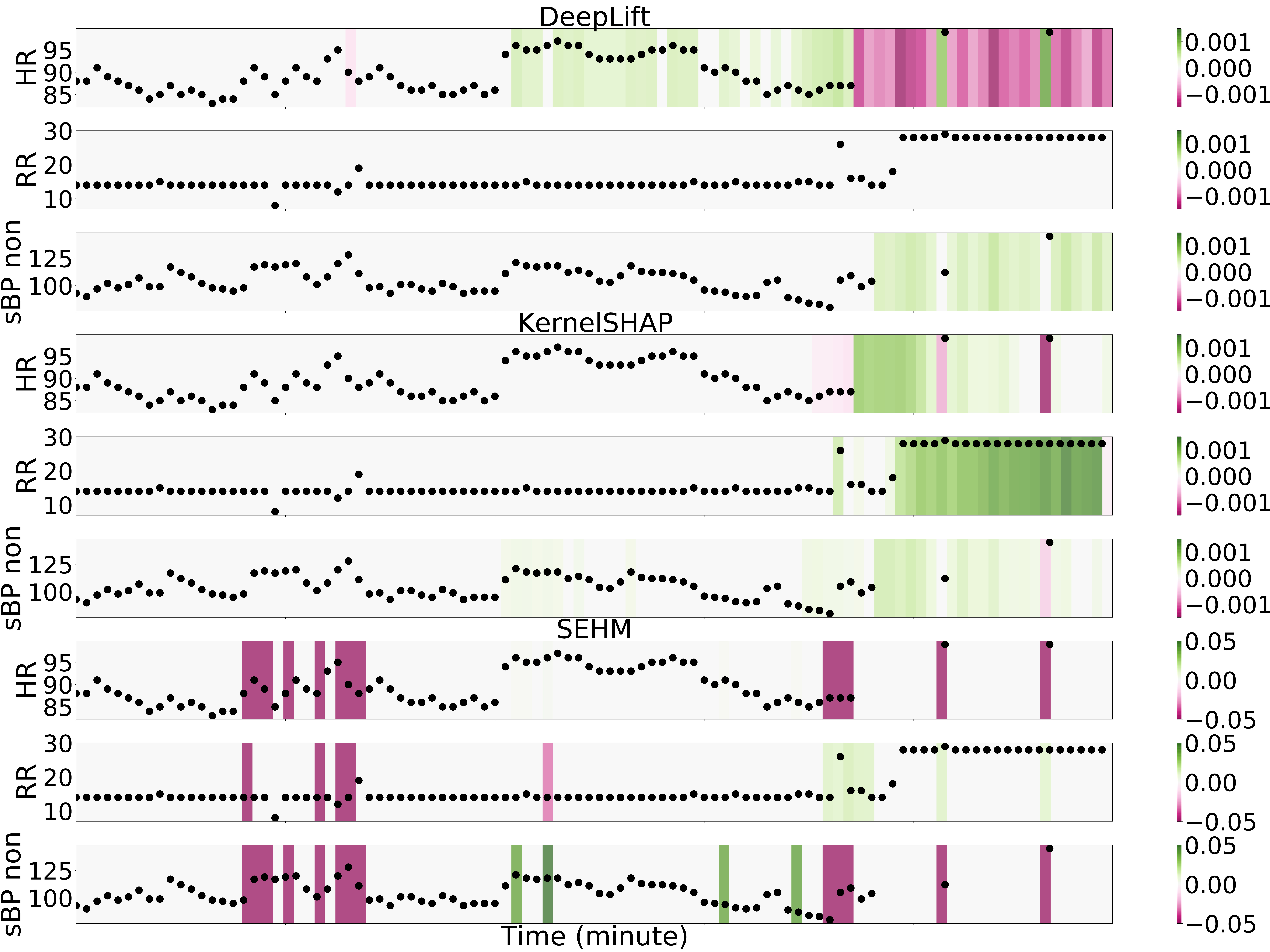}
        }
        
        \subfloat[Case B]{
        \label{fig:qual_vis_200}
		\includegraphics[scale=0.088]{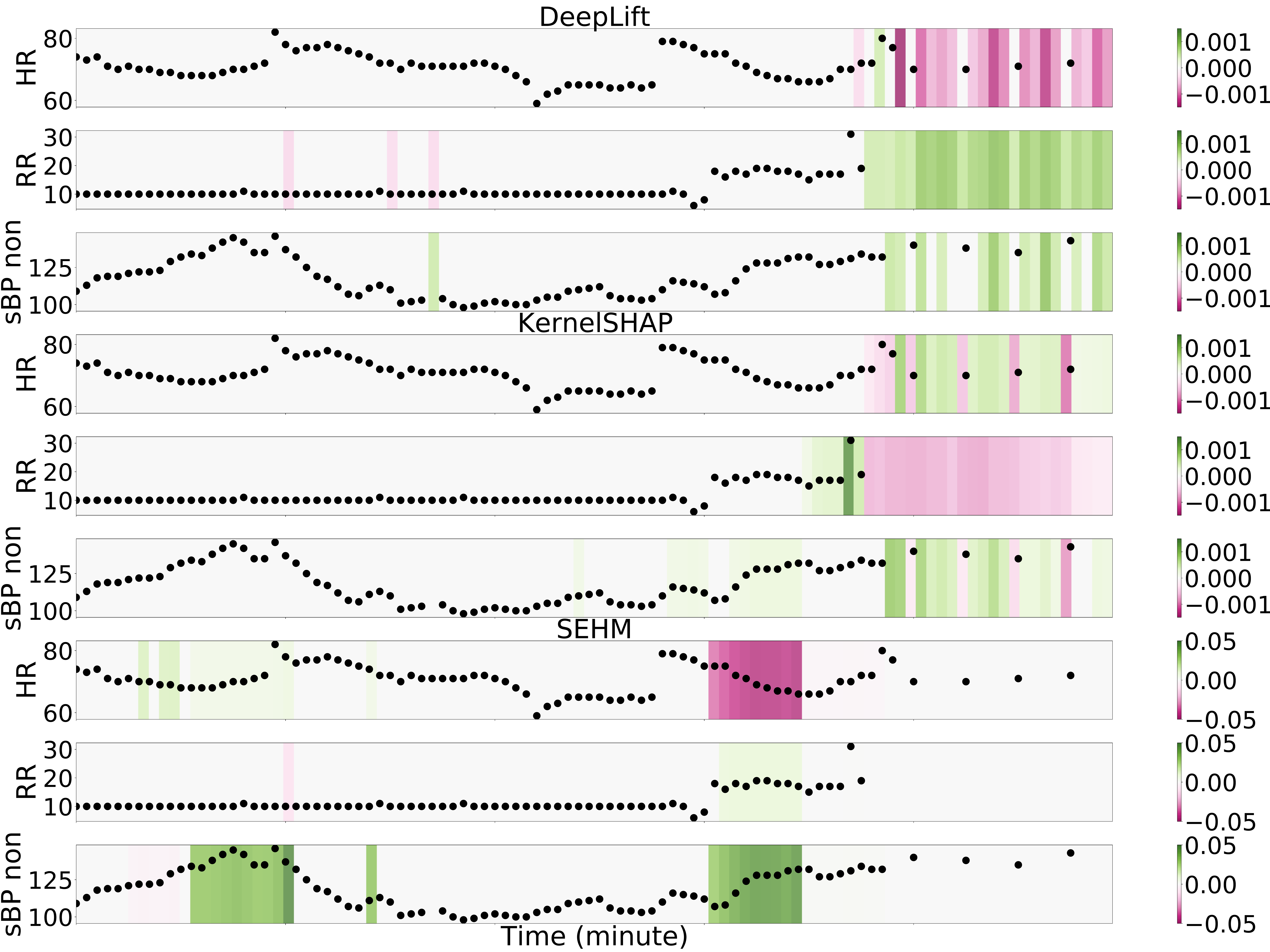}
        }
        \caption{Visualization of explanations generated for data points in the last 100 minutes of surgeries.}
\end{figure}

The visualizations shown in Figure~\ref{fig:qual_vis_86} and Figure~\ref{fig:qual_vis_200} are generated from the 100 consecutive minutes till the end of surgeries. We select 3 out of 56 intraoperative variables that are commonly available during the operation and intuitive to the readers. The selected variables include heart rate (HR), respiratory rate (RR) and non-intrusive blood pressure (sBP non). 

For the surgical case shown in Figure~\ref{fig:qual_vis_86}, SEHM marks the duration around the 20-th minute as highly important. Based on medical records, medications affecting blood pressure and heart rate were administrated to the patient at that time. However, both DeepLift and KernelSHAP miss the critical time associated with a medication event. In addition, SEHM identified a number of high values in the measurements with potential clinical significance. KernelSHAP attributes importance to the sequence of high RR values at the end of the case, even though the measurements are likely artifacts caused by an instrument issue.

Notably, both DeepLift and KernelSHAP assign high contributions to the end of surgery when few measurements were collected. The end-of-case sparse data issue is difficult for baseline methods to interpret because they tend to focus on missing time points, but missingness between observations is completely normal in that context. In contrast, SEHM avoids assigning importance to the end of surgery. This may be attributed to the design of SEHM that utilizes parameters learnt from global patterns during the training. Compared to DeepLift, SEHM may be more effective at learning from the global patterns. For many training cases, the end of surgery has sparse measurements that are not be correlated with the complication outcome.

In the second surgical case shown in Figure~\ref{fig:qual_vis_200}, SEHM highlights all the changes in four variables from the 60-th minute to the 70-th minute, which is during the patient's wake-up from the surgery. In contrast, the other two methods fail to capture this event. Moreover, SEHM is the only method that captures the increasing blood pressure from the 10-th minute to the 20-th minute by assigning high positive contributions. Again, SEHM is the only method that does not put too much emphasis on the end of surgery with sparse measurements. 

In general, the clinician's review of the two surgical case suggests the advantage of SEHM in identifying the variables and time windows in the input time series with potential clinical importance. 

\section{Conclusion}
This paper presents SEHM, a self-explaining hierarchical model specifically designed for intraoperative time series. SEHM integrates kernelized local attention and RNN to handle long and complex time series typical in intraoperative data. Furthermore, it provides end-to-end interpretability that identifies the input variables and the time windows in which the data are highly correlated to the final outcomes. Experiments on a real-world dataset demonstrate SEHM's superior performance in predicting postoperative complications when compared to state-of-the-art models. Furthermore, quantitative evaluation and case studies suggest the potential of SEHM in identifying clinical variables and time windows associated with predictions and important clinical events. An important direction for future work is to conduct a comprehensive evaluation of SEHM's impacts on clinicians' understanding of the predictions.

\section{Acknowledgement}
This work was supported, in part, by the Fullgraf Foundation.
\bibliography{references.bib} 
\bibliographystyle{ieeetr}
\end{document}